\documentclass[final,onefignum,onetabnum]{siamonline190516}

\usepackage{lipsum}
\usepackage{amsfonts}
\usepackage{graphicx}
\usepackage{epstopdf}
\usepackage{algorithmic}
\ifpdf
  \DeclareGraphicsExtensions{.eps,.pdf,.png,.jpg}
\else
  \DeclareGraphicsExtensions{.eps}
\fi

\usepackage{enumitem}
\setlist[enumerate]{leftmargin=.5in}
\setlist[itemize]{leftmargin=.5in}

\newsiamremark{remark}{Remark}
\newsiamremark{hypothesis}{Hypothesis}
\crefname{hypothesis}{Hypothesis}{Hypotheses}
\newsiamthm{claim}{Claim}

\headers{Correspondence of NNGP Kernel and the Mat{\'e}rn Kernel}{Amanda Muyskens, Benjamin W. Priest, Im\`ene R. Goumiri, and Michael D. Schneider}

\title{Correspondence of NNGP Kernel and the Mat{\'e}rn Kernel  \thanks{This work was performed under the auspices of the U.S. Department of Energy by Lawrence Livermore National Laboratory under Contract DE-AC52-07NA27344 with IM release number LLNL-TR-827489. Funding for this work was provided by LLNL Laboratory Directed Research and Development grant 19-SI-004.}}

\author{Amanda Muyskens\thanks{LLNL 
  (\email{muyskens1@llnl.gov}).}
\and 
Benjamin W. Priest \and 
Im{\`e}ne R. Goumiri \and 
Michael D. Schneider
}
\usepackage{amsopn}

\ifpdf
\hypersetup{
  pdftitle={Practical},
  pdfauthor={Amanda Muyskens, Benjamin W. Priest, Im\`ene Goumiri, and Michael Schneider}
}
\fi

\begin{document}

\maketitle

\begin{abstract}
	Kernels representing limiting cases of neural network architectures have recently gained popularity.
	However, the application and performance of these new kernels compared to existing options, such as the Mat{\'e}rn kernel, is not well studied.
	We take a practical approach to explore the neural network Gaussian process (NNGP) kernel and its application to data in Gaussian process regression.
	We first demonstrate the necessity of normalization to produce valid NNGP kernels and explore related numerical challenges.
	We further demonstrate that the predictions from this model are quite inflexible, and therefore do not vary much over the valid hyperparameter sets.
	We then demonstrate a surprising result that the predictions given from the NNGP kernel correspond closely to those given by the Mat{\'e}rn kernel under specific circumstances, which suggests a deep similarity between overparameterized deep neural networks and the Mat{\'e}rn kernel.
	Finally, we demonstrate the performance of the NNGP kernel as compared to the Mat{\'e}rn kernel on three benchmark data cases, and we conclude that for its flexibility and practical performance, the Mat{\'e}rn kernel is preferred to the novel NNGP in practical applications.
	
\end{abstract}
\begin{keywords}
  kernel, prediction, interpolation, Mat{\' e}rn covariance, neural network, Gaussian process regression 
\end{keywords}

\begin{AMS}
  68Q25, 68R10, 68U05
\end{AMS}

\section{Introduction}
\label{sec:intro}
Gaussian process regression is a convenient machine learning model that can flexibly approximate non-linear functions and provides principled uncertainty quantification of those non-linear predictions. 
A Gaussian process model is a continuous generalization of the normal distribution under the assumption that any finite set of data follows a joint multivariate normal distribution. 
Accordingly, the data model is fully specified by a mean and a covariance, which are often assumed to be parametric functions.
The covariance function is commonly called a kernel, and the mean is usually assumed to be zero without a loss of generality.
Defining new kernel functions is challenging in general since the forms must only produce positive definite covariance matrices to be a valid model.
Functions like radial basis functions (RBF) and the larger class of Mat{\'e}rn kernels are popular choices that are avantageous based on their universality and flexibility to interpolate general data. 
In 1996, Neal showed that properly initialized single-layer feedforward neural networks converge to particular Gaussian process kernels in the infinite width limit \cite{neal1996priors}. 
More recent work has generalized this result to deep neural networks (DNNs) \cite{lee2018deep, matthews2018gaussian}. 
Investigators have explored these dual representations in an attempt to learn about the behavior of different DNNs architectures \cite{jacot2018neural, garriga2018deep, lee2019wide, arora2019exact, yang2019tensor}. 
We focus here on the NNGP kernel (also called the conjugate kernel), which represents DNNs of a specified depth with infinitely wide, fully connected layers, with independent identically distributed normal weights where only the last layer is trained.

In this paper, we take a practical approach to the NNGP kernel and explore the implications of employing this kernel in model prediction.
Prediction from Gaussian process models (kriging) is derived from the conditional multivariate normal distribution and can be most simply described as a weighted average of the training observations. 
These weights are determined as a function of the kernel matrix, whose values are determined by the set of hyperparameters.
Therefore, we explore the effect of the hyperparameters of these kernels by computing and comparing their kriging weights directly. 
We compare kriging weights from the NNGP to those derived from the classical Mat{\'e}rn covariance function. 
We demonstrate a surprising practical equivalence in prediction between these NNGP kernels and the Mat{\'e}rn covariance kernel, and demonstrate the accuracy of these kernels on three example benchmark datasets.

\section{Background: Gaussian process Regression}
\label{sec:background}

We assume a linear model for a univariate response vector $y$ observed at training locations $X_\textrm{train} = [\mathbf{x}_1^T, \dots, \mathbf{x}_n^T]^T$ to be defined as $y=(y(x_1), y(x_2),...,y(x_n))^T$ for $n$ training observations.

We will assume that $\mathbf{f} \in \mathbb{R}^{n}$ constitute evaluations of a continuous, surrogate discrimination function $f_\theta: \mathbb{R}^L \rightarrow \mathbb{R}$ on $X_\textrm{train}$. 
We further assume that $\mathbf{y} \in \mathbb{R}^n$ are the ``observed''  realizations of $f_\theta$ on $X_\textrm{train}$ perturbed by homoscedastic Gaussian noise $\boldsymbol{\epsilon}$.
We seek to interpolate $f_\theta$'s response $\mathbf{f}_* \in \mathbb{R}^m$ to the unknown testing data $X^*_\textrm{test} = [\mathbf{x}^{*T}_1, \dots, \mathbf{x}^{*T}_m]^T$.

The GP assumption on $f_\theta$ amounts to the assertion that $f_\theta \sim \mathcal{GP}(\mathbf{0}, k_\theta(\cdot, \cdot))$, where $k_\theta$ is a positive semidefinite kernel function with parameters $\theta$.
$f_\theta \sim \mathcal{GP}(\mathbf{0}, k_\theta(\cdot, \cdot))$ imposes the following Bayesian prior model on $\mathbf{f}$, the true evaluations of $f$ on $X_\textrm{train}$:
\begin{align}
	\label{eq:prior_distribution}
	\begin{split}
		\frac{\mathbf{y}}{\sigma} &= \mathbf{f} + \boldsymbol{\epsilon}, \\
		\mathbf{f} &= [f_\theta(\mathbf{x}_1), \dots, f_\theta(\mathbf{x}_n)]^\top \sim \mathcal{N}(\mathbf{0}, K_\mathbf{ff}), \\
		\boldsymbol{\epsilon} &\sim \mathcal{N}(0, \tau^2 I_n).
	\end{split}
\end{align}
Here $K_\mathbf{ff}$ is an $n \times n$ positive definite covariance matrix on the training data whose $(i, j)$th element is $k_\theta(\mathbf{x}_i, \mathbf{x}_j)$, and $\tau^2$ is the variance of the unbiased homoscedastic noise.
The definition of GP regression then specifies that the joint distribution of all training and testing responses $\mathbf{y}$ and $\mathbf{f}^*$ is given by
\begin{equation}
	\label{eq:joint_distribution}
	\begin{bmatrix} \mathbf{y} \\ \mathbf{f}_*
	\end{bmatrix}
	= \mathcal{N} \left ( 0, \sigma^2
	\begin{bmatrix}
		K_\mathbf{ff} + \tau^2 I_n & K_\mathbf{f*} \\
		K_\mathbf{*f} & K_{**}
	\end{bmatrix}
	\right ).
\end{equation}
Here $K_\mathbf{f*} = K^\top_\mathbf{*f}$ is the cross-covariance matrix between the training and testing data;
that is, the $(i,j)$th element of $K_{f*}$ is $k(\mathbf{x}_i, \mathbf{x}^*_j)$.
Similarly, $K_\mathbf{**}$ is the covariance matrix of the testing data, and has $(i, j)$th element $k_\theta(\mathbf{x}^*_i, \mathbf{x}^*_j)$.
Finally, we are able to compute the posterior distribution of the testing response $\mathbf{f}^*$ on $X^*_\textrm{test}$ as
\begin{align}
	\label{eq:posterior_distribution}
	\begin{split}
		\mathbf{f}^* \mid X_\textrm{train}, X^*_\textrm{test}, \mathbf{y} &\sim \mathcal{N}(\bar{\mathbf{f}}^*, \sigma^2 C), \\
		\bar{\mathbf{f}}^* &\equiv K_\mathbf{*f} (K_\mathbf{ff} + \tau^2 I_n)^{-1} \mathbf{y}, \\
		C &\equiv K_{**} - K_\mathbf{*f} (K_\mathbf{ff} + \tau^2 I_n)^{-1} K_\mathbf{f*}.
	\end{split}
\end{align}

The quantity we refer to as the kringing weights ($H$) is the matrix (or vector if $X^*_\textrm{test}$ is one location) applied to the data vector $\mathbf{y}$ in order to obtain the predictions as
\begin{align}
	H = K_\mathbf{*f} (K_\mathbf{ff} + \tau^2 I_n)^{-1} .
\end{align}

\subsection{NNGP Kernel}
\label{sec:nngpkernel}

A fully-connected DNN with $M$ hidden layers and widths $\{ n^\ell \}^M_{\ell = 0}$ has parameters consisting of weight matrices $\left \{W^\ell \in \mathbb{R}^{n^\ell \times n^{\ell - 1}} \right \}^M_{\ell = 1}$ and biases $\left \{\mathbf{b}^\ell \in \mathbb{R}^\ell \right \}^M_{\ell = 1}$.
We initialize the weights and biases of our hypothetical DNN with i.i.d. $\mathcal{N}(0,1)$ variables.
We use hyperparameters $\sigma_a$ and $\sigma_b$ (effectively, variance priors for the weight and bias variables) to modify the variance of these parameter initializations.
The output of such a DNN on input $\mathbf{x}$ is $\mathbf{h}^M(\mathbf{x})$, computed recursively as
\begin{align} \label{eq:mlp}
	\begin{split}
		\mathbf{h}^1(\mathbf{x}) &= \frac{\sigma_a}{\sqrt{n^0}} W^1\mathbf{x} + \sigma_b b^1, \\
		\mathbf{h}^\ell(\mathbf{x}) &= \frac{\sigma_a}{\sqrt{n^{\ell-1}}} W^\ell \phi \left (\mathbf{h}^{\ell - 1}(\mathbf{x}) \right) + \sigma_b b^\ell,
	\end{split}
\end{align}
where here $\phi$ is an element-wise \emph{activation function}.
Translating the action of a DNN layer in the infinite width limit to kernel form requires obtaining a dual form of the nonlinearity $\phi$ given positive definite kernel matrix $K$.
Several popular activation functions have known dual forms (see \cite{daniely2016toward} for some examples).
We will follow other research on this topic by focusing on the popular rectified linear unit (ReLU) activation, which fortunately has a known analytic dual form given by
\begin{align} \label{eq:Vphi}
	\begin{split}
		V_{\phi_{\text{ReLU}}} (K) (\mathbf{x}, \mathbf{x}^\prime)
		&= \mbox{$\frac{\sqrt{K(\mathbf{x}, \mathbf{x}) K(\mathbf{x}^\prime, \mathbf{x}^\prime)}}{2 \pi} \left ( \sin c + ( \pi - c) \cos c) \right )$}  \\
		V_{\phi_{\text{ReLU}}^\prime}  (K) (\mathbf{x}, \mathbf{x}^\prime)
		&= \frac{1}{2 \pi} ( \pi - c)  \\
		c
		&= \arccos \left ( \frac{K(\mathbf{x}, \mathbf{x}^\prime)}{\sqrt{K(\mathbf{x}, \mathbf{x}) K(\mathbf{x}^\prime, \mathbf{x}^\prime)}} \right ).
	\end{split}
\end{align}
Using this dual activation and the notation of Equation~\eqref{eq:mlp} and following the formulation of \cite{yang2019fine}, we can express the NNGP recursively as
\begin{align} \label{eq:ck}
	\begin{split}
		\Sigma^1(\mathbf{x}, \mathbf{x}^\prime)
		&= \frac{\sigma^2_a}{n^0} \langle \mathbf{x}, \mathbf{x}^\prime \rangle + \sigma^2_b, \\
		\Sigma^\ell(\mathbf{x}, \mathbf{x}^\prime)
		&= \sigma^2_a V_{\phi_\textrm{ReLU}}(\Sigma^{\ell -1}) (\mathbf{x}, \mathbf{x}^\prime) + \sigma^2_b, \\
		k_{(M, \sigma_a, \sigma_b)}(\mathbf{x}, \mathbf{x}^\prime) 
		&= \Sigma^M(\mathbf{x}, \mathbf{x}^\prime).
	\end{split}
\end{align}



\subsection{Mat{\'e}rn Kernel}
The posterior distribution given in Equation~\eqref{eq:posterior_distribution} depends on the choice of kernel function.
We will analyze the Mat\'ern kernel, which is a stationary and isotropic kernel that is commonly used in the spatial statistics GP literature due to its flexibility and favorable properties \cite{stein2012interpolation}.
A general expression for the kernel is
\begin{equation}
	k_{(\sigma^2, \nu, \ell)}(\mathbf{x},\mathbf{x^\prime}) =  \sigma^2 \frac{2^{1-\nu}}{\Gamma(\nu)}
	{\left(
		\sqrt{2\nu} \frac{{\| \mathbf{x} - \mathbf{x^\prime} \|}_2^2}{\ell}
		\right)}^\nu
	B_\nu{\left(
		\sqrt{2\nu} \frac{{\| \mathbf{x} - \mathbf{x^\prime} \|}_2^2}{\ell}
		\right)},
	\label{eq:matern}
\end{equation}
where $\nu>0$ is a smoothness parameter, $\ell>0$ is a correlation-length scale hyperparameter, $\sigma^2>0$ is a scale parameter, $\Gamma$ is the Gamma function, and $B_\nu(\cdot)$ is a modified Bessel function of the second kind.
Note that as $\nu \to \infty$, the Mat{\'e}rn kernel converges pointwise to the popular radial basis function (RBF) kernel.
We will compare the predictions of the Mat{\'e}rn and NNGP kernels because the former is considered to be state-of-the art in terms of flexibility and general applicability to realistic data.

\section{Application of the NNGP Kernel}
\label{sec:nngp}
In this section, we discuss the practical application of the NNGP kernel for Gaussian prcoess regression. 
We examine the common normalization scheme conventional in applying the NNGP as well as limit parameters $\sigma_a$ and $\sigma_b$ to sets that give valid covariance matrices. 
Next, we explore kriging weights given by the NNGP kernel under those valid parameter sets and compare the results to the kriging weights of the common Mat{\'e}rn kernel.

\subsection{Normalization}
It is conventional (see \cite{lee2018deep, matthews2018gaussian}) to embed the data on the unit hypershere prior to applying the NNGP kernel.
Formally, the unit hypersphere embedding we employ is as follows.
Much of the published research analyzing the NNGP involves image data, where such normalization amounts to normalizing the image intensity across all images.
We will employ a slightly different normalization method to accomodate lower dimensional data.
Let $X_i$ be the $i^{th}$ column of the raw training input data.
Then create training data matrix $X^{\star}$ to have twice as many dimensions so that each column in $X$ is two in $X^{\star}$ as
\begin{equation}
X_{2i+1,2i+2}^{\star} = \left[cos(X_i \pi), sin(X_i \pi)\right].
\end{equation}
Finally, these columns are additionally normalized by their L2-norms so
\begin{equation}
	X_{i} = \frac{X_{i}^{\star}}{||X_{i}^{\star} ||_2}.
\end{equation}
Finally, define $X_{train}$ and $X_{test}^{\star}$ to be the matrices containing these columns normalized by these two procedures.

\begin{figure}
	\centering
	\includegraphics[scale=.4]{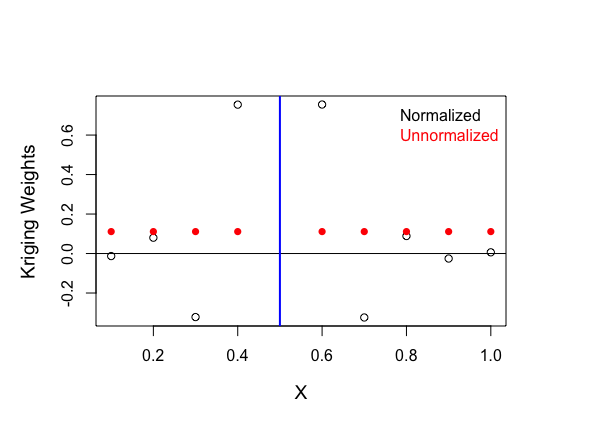}
	\caption{Comparison of kriging weights from a 1-dimensional Gaussian process regression compared on data normalized to the unit hypershpere and unnormalized. }
	\label{fig:norm}
\end{figure}

Figure \ref{fig:norm} demonstrates the difference in kriging weight for prediction at the vertical line produced from a 1-dimensional example for normalized and unormalized implementation of the NNGP kernel. 
Note that if the data is not normalized, the prediction in this case is approximately the mean of the training data.

\subsection{Valid Hyperparameter Sets}
\begin{figure}
	\centering
	\includegraphics[scale=.4]{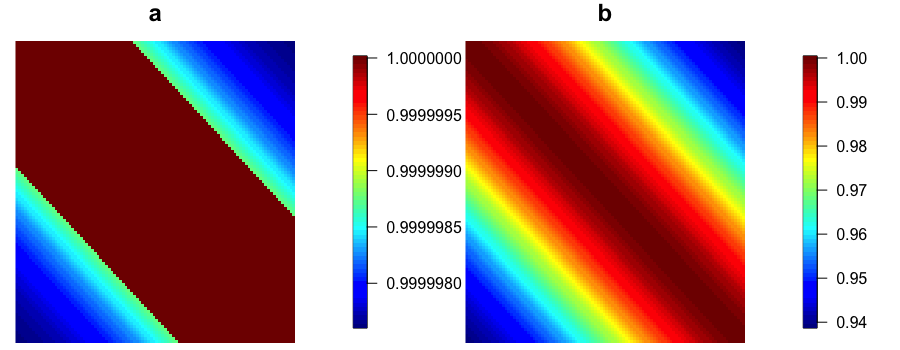}
	\caption{a. Invalid NNGP kernel b. Valid NNGP Kernel}
	\label{fig:kernel}
\end{figure}

Once the data is normalized as described, we examine ranges of the NNGP kernel hyperparameters $\sigma_a$ and $\sigma_b$. 
Assuming that $X_{train}$ is a 1-dimensional grid on the interval $(0,1]$, we formed the NNGP kernel $k_{(M, \sigma_a, \sigma_b)}$ for hyperparameters on a $20 \times 20$ grid for  $\sigma_a, \sigma_b \in [0.1, 2.0]$ for various depths. 
Some kernels generated were numerically invalid covaraince matrices because they were not positive definite and had off-diagonal covariance entries equal to the diagonal entry.
This produces "flat" regions in the matrix where the correlation is numerically equal to 1 for non-zero distances.
Examples of an invalid kernel matrix and a valid kernel matrix are in Figure \ref{fig:kernel}.
In the invalid kernel matrix, the correlation over the entire domain is extremely high, and although there is a wider range of correlation in the valid kernel example, data across the domain being correlated as a minimum of 0.94 is still unreasonably high for most realistic data. 

\begin{figure}
	\centering
	\includegraphics[scale=.7]{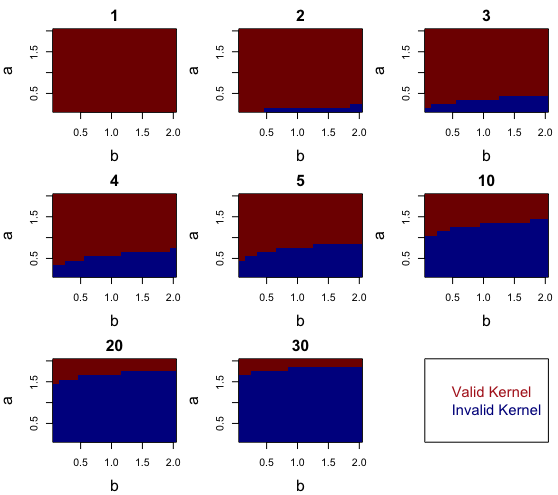}
	\caption{Valid NNGP hyperparameter sets that produce positive definite kernel matrices for various depths for NNGP hyperparameters $\sigma_a$ and $\sigma_b$.}
	\label{fig:range}
\end{figure}

\begin{figure}
	\centering
	\includegraphics[scale=.4]{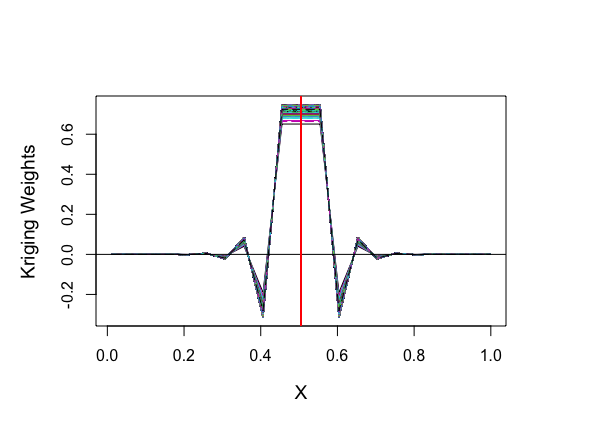}
	\caption{Variation of kriging weights over valid NNGP hyperparameter sets and depths.}
	\label{fig:spread}
\end{figure}

Figure \ref{fig:range} evaluates which hyperparameter sets produce valid kernel matrices.
As the depth of the corresponding NN increases, the region of hyperparameter space that produces degenerate matrices increases.
Finally, Figure \ref{fig:spread} demonstrates the variation in the kriging weights for all considered depths and parameters in Figure \ref{fig:range}.
Even in these diverse parameter settings, the kriging weights have very little variation and therefore will produce extremely similar predictions.

\subsection{Kriging Weights Correspondence to Mat{\'e}rn Kernel}

\begin{figure}
	\centering
	\includegraphics[scale=.4]{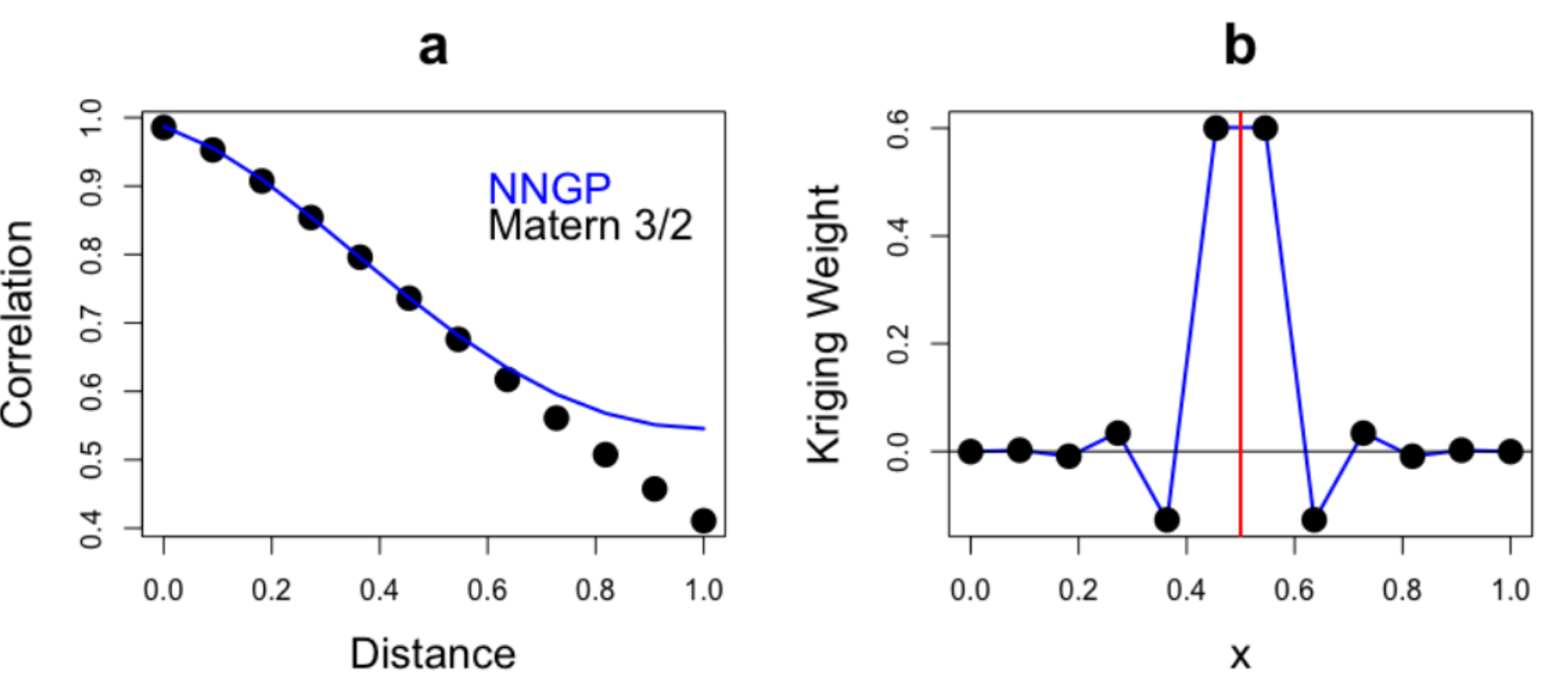}
	\caption{Comparison of correlation functions (a) and kriging weights (b) of NNGP (blue) and Mat{\'e}rn $\frac{3}{2}$ (black) kernels in 1-D for prediction at $x=0.5$ (red). }
	\label{fig:corr}
\end{figure}

In this section we demonstrate the surprising correspondence between kriging weights produced from the NNGP kernel and the Mat{\'e}rn kernel with smoothness parameter $\nu=\frac{3}{2}$.
Although the kriging weights ($H$) from the two kernels are extremely similar under certain conditions, this correspondence is not numerically exact, and actually the correlation function ($k_\theta$) that produces the kernel matrices do not themselves correspond over the entire data range. 
Figure \ref{fig:corr} provides an example of this phenomenon.
Figure \ref{fig:corr}a shows that the NNGP and Mat{\'e}rn and NNGP correlation functions diverge as distance increases, and yet Figure \ref{fig:corr}b shows that the resulting kriging weights agree.
There is significant agreement in the correlation function for the normalized distances less than 0.5.

%

Computationally efficient Gaussian process methods including \cite{gramacy2015local} and \cite{datta2016hierarchical} have exploited locality-induced sparsity in order to gain their computational efficiency.
Further, high-frequency trends (local correlations) are more important to accurate prediction than low-frequency trends (long-scale correlations) \cite{stein2012interpolation}.
Our findings extend this knowledge, by demonstrating similar predictions are produced with only local correlation similarity.

\begin{figure}
	\centering
	\includegraphics[scale=.6]{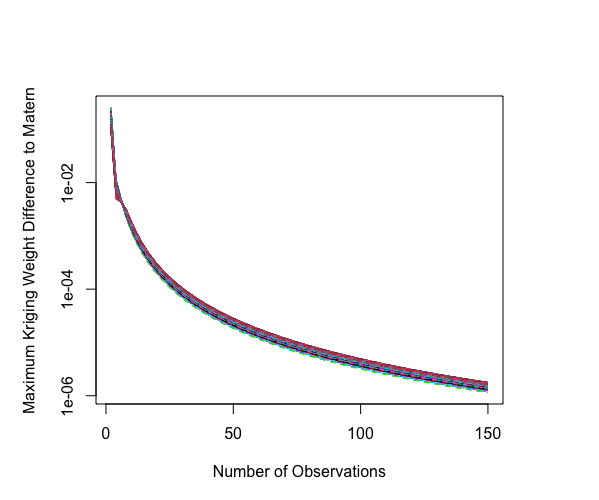}
	\caption{Maximum absolute difference between kriging weights produced from the Mat{\'e}rn $\nu=\frac{3}{2}$ and those from the NNGP kernel in a grid in 1-D.}
	\label{fig:grid}
\end{figure}

\begin{figure}
	\centering
	\includegraphics[scale=.6]{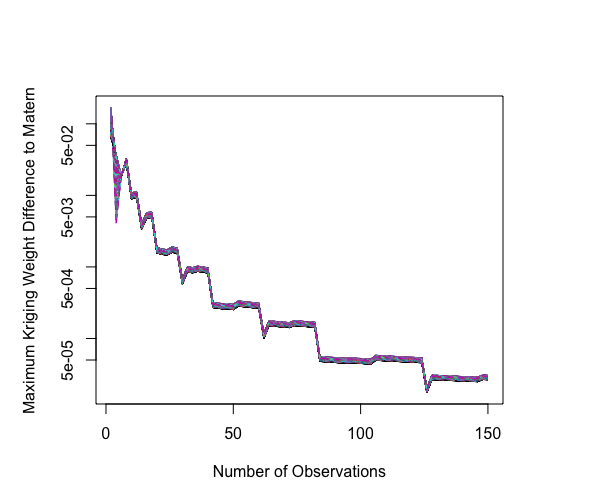}
	\caption{Maximum absolute difference between kriging weights produced from the Mat{\'e}rn $\nu=\frac{3}{2}$ and those from the NNGP kernel sampled according to the quasi-random sobol sequence in 1-D.}
	\label{fig:sobol}
\end{figure}

As the number of observations increases, the similarity between the kringing weights also increases.
As there are more observations normalized to $[0,1]$, this implies that there are more observations with corresponding correlations.
Using data generated both on grid (Figure \ref{fig:grid}), and generated using a quasi-random sobol sequence (Figure \ref{fig:sobol}), we demonstrate the maximum absolute difference between the Mat{\'e}rn kriging weights ($\nu = \frac{3}{2})$ and those from the NNGP.
In both cases, with data sizes at about 150, the largest difference in kriging weights is less than 0.00005.

Although we demonstrate the similarity between the kriging weights only from data in 1-dimension, a similar correspondence can be seen in higher dimensional data.
The next section shows the practicality of this correspondence in higher dimensional examples.
Also, \cite{muyskens2021ident} demonstrates that the Mat{\'e}rn kriging weights are only dependent on the $\nu$ parameter when the nugget parameter $\tau^2$ is small (very close to 0).
However, when $\tau^2$ is large, the kriging weights also depend on the range parameter $\rho$.
We omit the results here for similcity, but if $\tau^2$ is large, there is still a correspondence in predictions, but $\rho$ must also be appropriately selected.
In summary, under large and densely-sampled data, predictions from the NNGP kernel are practically similar to those from the Mat{\'e}rn kernel with $\nu = \frac{3}{2}$.
The next section will demonstrate similar these predictions are in several benchmark data cases.

\section{Numerical Demonstrations}
\label{sec:num}
In this section, we numerically demonstrate the importance of the conclusions of the previous section by applying Gaussian process regression with the NNGP kernel on several benchmark datasets. 

Throughout this section, we will compare Gaussian processes with three kernel ($k_\theta$) assumptions.
\begin{enumerate}
	\item NNGP kernel with $\sigma_a, \sigma_b$ varied, depth$=2$
	\item Mat{\'e}rn kernel with $\nu=\frac{3}{2}$ and $\rho=1$
	\item Mat{\'e}rn kernel with $\nu, \rho$ varied
\end{enumerate}
Note that we compare to the NNGP kernel with low depth so that the entire range of $\sigma_a$ and $\sigma_b$ yields a valid covariance model, but the results should not be very different if another depth was selected since the kriging weights, and therefore predictions, from various depths are similar (Figure \ref{fig:spread}).
Also, we do not perform a grid search to optimize parameters on a training subset of the data, but instead report just the best performance of the testing data to demonstrate the best performance from each kernel.
To compare the results of these three models, we consider several summary statistics.
First, best accuracy of the models is compared via root mean squared error (minRMSE) of the model predictions. 
\begin{equation}
	minRMSE = \min_\theta\sqrt{\frac{1}{m}\sum_{i=1}^{m}(\hat{Y}_i-Y_i)^2},
\end{equation}
where $\hat{Y}_i=H_{i\theta} Y$.
Similarly, we report the worst accuracy of each kernel (maxRMSE) as
\begin{equation}
	maxRMSE = \max_\theta\sqrt{\frac{1}{m}\sum_{i=1}^{m}(\hat{Y}_i-Y_i)^2}.
\end{equation}
Next, we consider several statistics describing the differences in kriging weights between the best NNGP kernel and each kernel. 
Formally, define the best performing NNGP kernel parameter set to have corresponding kriging weights matrix $H_{NNGP}$, which is dimension $m \times n$ when we utilize $n$ training data observations in order to predict at $m$ testing locations.
Define $\tilde{\theta}$ to be the parameter set for each kernel that minimizes the maximum absolute difference in the kriging weights as compared to $H_{NNGP}$.
Then for each kernel type we report the maximum (maxdiff), minimum (mindiff), mean (meandiff), and standard deviation (sddiff) of the difference in the kriging weights as compared to those from the best-performing NNGP kernel.
\begin{equation}
\text{maxdiff}= \max_{n,m}|H_{\tilde{\theta}}-H_{NNGP}|
\end{equation}
\begin{equation}
	\text{mindiff}=min_{n,m}|H_{\tilde{\theta}}-H_{NNGP}|
\end{equation}
\begin{equation}
	\text{meandiff}= \frac{1}{nm} \sum_{i=1}^{nm}|H_{\tilde{\theta}}-H_{NNGP}|
\end{equation}
\begin{equation}
	\text{sddiff} = \frac{1}{nm-1} \sqrt{\sum_{i=1}^{nm}|H_{\tilde{\theta}}-H_{NNGP}|^2}
\end{equation}
Finally, we look at the distribution of the differences across $\theta$ values for each kernel form. 
For each parameter set $\theta$, we compute the maximum difference in the kriging weights produced as related to those from the mean kriging weight $\bar{H}$. 
Then we report summary statistics of that maximum difference over the grid search of values tested. 
These summaries demonstrate the variance in the kriging weights over various sets tests, ie a more flexible kernel form.
Define $n_\theta$ to be the number of $\theta$ parameter sets for each kernel type.
Then we report similar distributional statistics of the kriging weights defined as
\begin{equation}
	\text{maxkw}= \max_{\theta}max|H_{\theta}-\bar{H}|
\end{equation}
\begin{equation}
	\text{minkw}=\min_{\theta}max|H_{\theta}-\bar{H}|
\end{equation}
\begin{equation}
	\text{meankw}= \frac{1}{nm} \sum_{i=1}^{nm}max|H_{\theta}-\bar{H}|
\end{equation}
\begin{equation}
	\text{sdkw} = \frac{1}{n_\theta-1} \sqrt{\sum_{i=1}^{n_\theta}max(H_{\theta}-H_{NNGP})^2}
\end{equation}
The standard deviation over 100 simulation iterations of each statistic is also reported.
%

\subsection{Friedman Function Data}
\begin{figure}
	\centering
	\includegraphics[scale=.55]{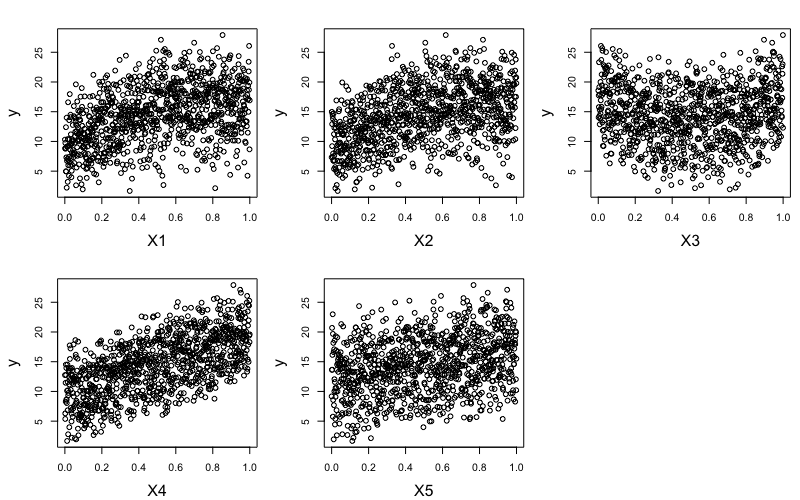}
	\caption{ Marginal plots of the responses from the Friedman function against each input variable. }
	\label{fig:friedman}
\end{figure}

Data in this comparsion is simulated from the common surrogate model Friedman function defined as
\begin{equation}
	Y_i = 10sin(\pi x_{1i}*x_{2i}) + 20(x_{3i}-0.5)^2+ 10x_{4i}+ 5x_{5i} + \epsilon_i,
\end{equation}
where $\epsilon_i\sim N(0, 1)$ for $x_i \in [0,1]$ for $i=1,2,3,...5$.
We sample 500 training locations according to a random Latin Hypercube design, and model the responses as follows:
\begin{equation}
	Y_i \sim GP(X_i\beta, k_\theta(x_i,x_i)),
\end{equation}
where $k_\theta$ is one of the three outline kernel models in this section.
Example realizations from this function plotted by the input variables can be seen in Figure \ref{fig:friedman}

\subsection{MODIS Satellite Dataset}
\begin{figure}
	\centering
	\includegraphics[scale=.45]{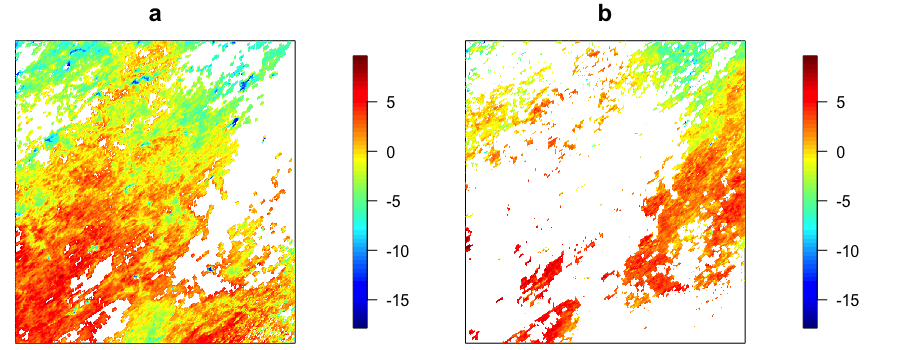}
	\caption{a. Full training data
		b. Full testing data \label{fig:heaton}. We subset 500 random samples from each of these for data in a single simulation iteration. }
\end{figure}

Our data are sourced from the numerical comparisons in \cite{heaton2019case} and can be downloaded at https://github.com/finnlindgren/heatoncomparison.
It is land surface temperature data from latitude ranging from 34.29519 to 37.06811 and longitudes  from -95.91153 to -91.28381 collected on August 4, 2016.
The 148,309 observations were collected. on a paritially missing $500\times 300$ grid.
The cloud cover from August 6, 2016 was used in order to develop a realistic testing/training data split.
There are 42,740 testing obbservations and 105,569 training observations with the pattern as seen in Figure~\ref{fig:heaton}.
Since our comparison is meant to compare kernel functions, we fist subtract the Gaussian filtering mean from the data as in \cite{muyskens2021muygps}.
Then our comparison fits the residuals from this analysis as the ground truth.
Both the training data and testing data are randomly downsampled so that kriging estimators can be implemented in typical laptop constraints.
We randomly sample 500 training and 500 testing locations.
The data is formally modeled
\begin{equation}
	Y_i \sim GP(\mu_i, k_\theta(x_i,x_i)),
\end{equation}
where $k_\theta$ is one of the three outline kernel models in this section, and $\mu_i$ is the moving window mean in \cite{muyskens2021muygps}.
We compare the performance of the aformentioned three models in Table \ref{table:1}.

\subsection{Borehole Function}
The borehole function (\cite{worley1987deterministic}) is a common benchmark function used to demonstrate computer simulation modeling (\cite{cole2021locally, katzfuss2020scaled}).
It is meant to model the flow of water through a borehole, and includes parameters such as the radius of the borehole in meters ($r_w$), the hydraulic conductivity of the borehole in m/yr ($K_w$) and 6 other parmeters that effect the flow.
This 8-dimensional funtion is defined as:
\begin{equation}
	y = \frac{2\pi T_u [H_u-H_l]}{\log(\frac{r}{r_w})[1 + \frac{2LT_u}{\log(r/r_w) r_w^2 K_w} +\frac{T_u}{T_l}]}
\end{equation}
where $r_w \in [0.05, 0.15]$,
$H_u \in [990, 1100]$,
$r \in [100, 5000]$,
$H_l \in [700, 820]$,
$T_u \in [63070. 115600]$,
$L \in [1120, 1680]$,
$T_l \in [63.1, 116]$,
$K_w \in [9855, 12045]$.

\begin{figure}
	\centering
	\includegraphics[scale=.4]{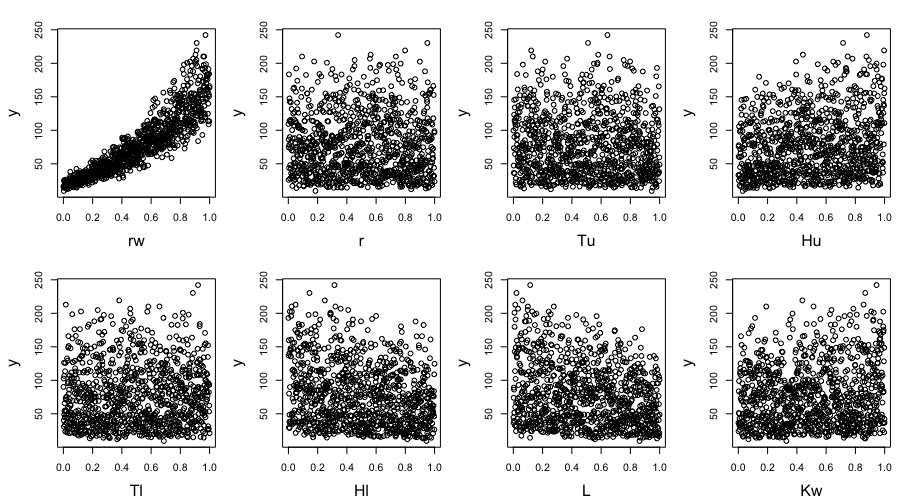}
	\caption{Marginal plots of the responses from the borehole function against each input variable.}
	\label{fig:borehole}
\end{figure}

We formally model this function as we modeled the Friedman function with a linear mean.
Formally,
\begin{equation}
	Y_i \sim GP(X_i\beta, k_\theta(x_i,x_i)),
\end{equation}
where $k_\theta$ is one of the three outline kernel models in this section.
Example realizations from this function plotted by the input variables can be seen in Figure~\ref{fig:borehole}.

\subsection{Results}
\begin{figure}
	\centering
	\includegraphics[scale=.4]{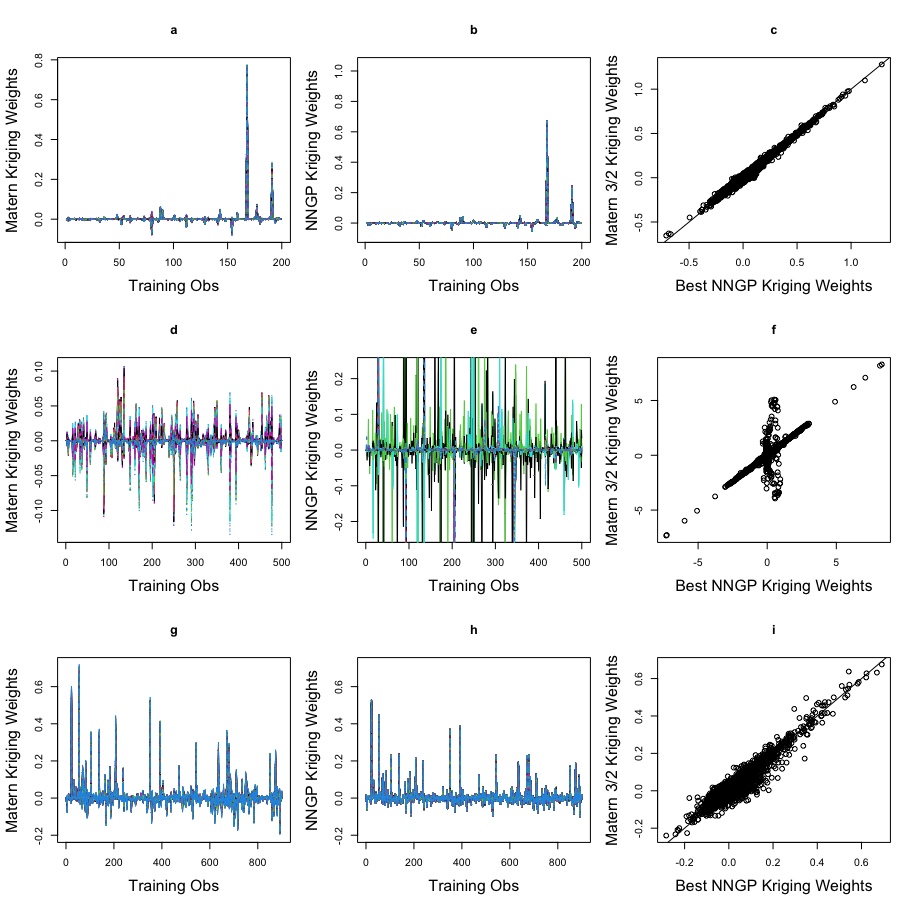}
	\caption{Plots of kriging weights of the Mat\'ern (a,d,g), NNGP (b, e, h), and a scatterplot comparison of the most performant NNGP kriging weights vs. the kriging weights of the Mat\'ern $\nu=\frac{3}{2}$ (c, f, i). These results are for one iteration simulation for the Friedman function (a,b,c), MODIS satellite data (d,e,f), and borehole function (g,h,i).}
	\label{fig:res}
\end{figure}

Table \ref{table:1} summarizes 100 iterations of these benchmark examples as outlined. 
The NNGP and Mat{\'e}rn $\nu=\frac{3}{2}$ kernels perform relatively similarly, but the Mat{\'e}rn model with varied parameters has the smallest RMSE in all cases by a wide margin. 
Note that although we have demonstrated the convergence of the kriging weights under ideal conditions, this ``correspondence'' is not absolute in the sense the compared kernels yield similar but not identical predictions. 
In terms of maxRMSE, the Mat{\'e}rn $\nu=\frac{3}{2}$ produces the best worst-case prediction error so if one were to implement a kernel without optimization for testing purposes, this would be the best kernel selection in general. 
In the Friedman function and the borehole function data examples, our results are as expected where the kriging weights from the  Mat{\'e}rn kernel are generally close to those of the NNGP kernel.
However, we see a large deviance in the NNGP kriging weights in some iterations.
When a matrix is close to singular (as some valid NNGP kernels are), when data is observed very closely, numerical difficulties can cause the kriging weights to become artificially large.
This is clearly the case in the MODIS data example, and ultimately results in extremely poor performance for a few of the iterations.
This numerical challenge is seen here in the NNGP kernel, but not the Mat{\'e}rn kernel results.
An example iteration of the kriging weights from the various kernels are compared in Figure \ref{fig:res}.

\begin{table}[ht]
	\centering
	\begin{tabular}{rcrrr}
	Statistic & Data & NNGP&Mat{\'e}rn $\nu=\frac{3}{2}$ & Mat{\'e}rn  \\ [0.5ex] 
  	\hline
 	&Friedman& 0.054 (0.005)& 0.026 (0.007) & \textbf{0.005}(0.002) \\ 
    minRMSE& MODIS& 4.294 (1.534)& 4.892 (2.288)& \textbf{2.216} (0.126) \\ 
  	& Borehole& 1.520  (0.210)& 1.288 (0.394) & \textbf{0.406} (0.144)\\ 
\hline

 & Friedman& 0.063 (0.007)& 0.026 (0.007)& 4.198 (0.231)\\ 
  maxRMSE & MODIS& 4425.090(13059.737)& 4.892 (2.288)& 16.054 (11.756)\\ 
   & Borehole & 1.960 (0.354)& 1.288 (0.394)& 88.990 (5.556)\\
   
   \hline
    \hline
   
   &Friedman &0.000 (0.000)& 0.299 (0.450) & 0.225 (0.355) \\ 
 maxdiff& MODIS&0.000 (0.000)& 8.719 (10.611)& 7.659 (10.944) \\ 
 & Borehole&0.000 (0.000)& 0.209 (0.033)& 0.201 (0.030) \\ 
 
  \hline
 
 & Friedman&0.000 (0.000)& 1.2e-08 (1.3e-08 )&1.4e-08 (1.6e-08)\\
mindiff&MODIS&0.000 (0.000)& 1.8e-12 (1.8e-12)&1.9e-12 (2.6e-12)\\
&Borehole&0.000 (0.000)& 5.7e-07 (5.3e-07)&8.0e-07 (6.8e-07)\\

 \hline
 &Friedman&0.000 (0.000)& 0.003 (0.000) & 0.003 (0.001)\\ 
  meandiff&MODIS&0.000 (0.000) & 0.004 (0.005)& 0.011 (0.012)\\ 
  &Borehole &0.000 (0.000)& 0.011 (0.001)& 0.012 (0.001)\\ 
   \hline

&Friedman&0.000 (0.000)& 0.005 (0.001)& 0.006 (0.003) \\ 
  sddiff&MODIS&0.000 (0.000)& 0.072 (0.067)& 0.102 (0.117) \\ 
  &Borehole&0.000 (0.000)& 0.012 (0.001)& 0.013 (0.001)  \\ 
  
   \hline
    \hline
  
  &Friedman& 0.526 (1.202)& 0.000 (0.000)&1.289 (0.365)\\ 
 maxkw& MODIS & 116177.8(532722.5)&0.000 (0.000)& 30.469 (17.109)\\ 
 & Borehole& 0.015 (0.004)&0.000 (0.000) &0.649 (0.139)  \\

   \hline
 & Friedman&1.3e-06 (4.0e-07)&0.000 (0.000)& 2.6e-05 (5.6e-06)\\
minkw&MODIS&1.3e-06 (1.9e-06)&0.000 (0.000)&7.4e-08 (1.4e-08)\\
&Borehole&1.5e-05 (4.2e-06)&0.000 (0.000)& 4.7e-04 (8.0e-05)\\

 \hline

&Friedman& 0.000 (0.000)& 0.000 (0.000)&0.011 (0.000)\\ 
 meankw& MODIS& 3.079 (10.018)&0.000 (0.000) &0.056 (0.017)\\ 
 & Borehole & 0.001 (0.000)&0.000 (0.000)& 0.020 (0.001)\\ 
   \hline
   
 & Friedman& 0.002 (0.004)& 0.000 (0.000)& 0.030 (0.001) \\ 
  sdkw &MODIS& 379.826(1637.020)& 0.000 (0.000)& 0.480 (0.205) \\ 
  &Borehole& 0.001 (0.000)& 0.000 (0.000)& 0.030 (0.002) \\ 
	\end{tabular}
	\caption{Performance of various kernel models for the data examples described in Section \ref{sec:num}. Parthenses contain the standard deviation of the estimate over 100 random iterative draws. }
	\label{table:1}
\end{table}

\section{Discussion}
In this manuscript, we have demonstrated the practical considerations of utilization of the NNGP kernel for Gaussian process regression.
We have shown normalization to the unit hypersphere is necessary in order to obtain meaningful predictions.
We have also shown that there are NNGP parameter sets that return degenerate covariance matrices, and that the regions of parameter space that have this numerical issue tend to increase in size as depth increases. 
We have demonstrated that for useable parameter combinations, when the data is sufficiently large, predictions from the NNGP kernel are approximately equivalent to those from the Mat{\'e}rn $\nu=\frac{3}{2}$ model. 

We interpret this result to mean that theoretical well-converged neural networks in the infinite width limit are essentially Gaussian processes with the well-known Mat{\'e}rn kernel function.
Determining uncertainty quantification (UQ) is currently a major research focus in the study of DNNs and other machine learning methods \cite{kabir2018neural}.
As predictions from GP models and idealized DNNs under certain architectural assumptions are the same, further research could demonstrate whether the UQ from the equivalent GP model could be realistically be applied to trained DNNs. 

Further, it is generally accepted that DNNs demonstrate complex, non-local dependence in the data. 
However, the practical correspondence to the Mat{\'e}rn kernel challenges this idea, at least for the specific architecture the NNGP represents.
In \cite{muyskens2021muygps}, it is shown that the kriging weights from Gaussian process regression with a Mat{\'e}rn are sparse in distance to the prediction location, meaning they depend most largely on nearest neighbors for similar predictions.
Therefore, since the predictions from the NNGP are similar to those from the Mat{\'e}rn, in this case, they are also locally-dominated predictions.

Our work could be continued through the comparision of the performance of these kernels in higher dimensional cases. 
For example, \cite{muyskens2021star} utilize a PCA reduction and GP regression in order to perform image classification. 
Even with this data reduction, the dimension of data fit is 50, which is much higher than the examples we have considered in this manuscript.
This higher dimensional data is a more typical dataset for neural networks, and GP models are known to struggle in prediction in these cases. 

Additionally, NNGP is not the only GP kernel that has been identified corresponding to other neural network architectures. 
One example is the neural tangent kernel (NTK), which is similar in form to the NNGP kernel and is used in the study of training dynamics \cite{jacot2018neural}.
Furthermore, kernel correspondences to infinite width limits of other architecture types such as convolutional, recurrent, and graph neural networks have emerged in the literature \cite{garriga2018deep, yang2019tensor}.
Further studies could explore these other kernels, and evaluate if there exists prediction equivalence to the predictions from these models. 

In conclusion, the NNGP kernel allows us to study predictions from an infinitely wide neural network with the weights and biases as i.i.d. $N(0,1)$ variables.
This has allowed us to understand that predictions from this theoretical neural network .
Future research will need to extend our results to understand whether these conclusions generalize to more complex NN architectures. 
However, this kernel is more interesting in theory than in application.
It has numerical challenges where it produces non-valid kernel matrices, particularly in deeper architectures.
It under-performed in several benchmark datasets as compared to the more common Mat{\'e}rn kernel.
Therefore, in application for its flexibility and performance, the classic Mat{\'e}rn kernel appears to remain the most practical kernel model for application of Gaussian process models to data of the types explored in this study.

\appendix

\section*{Acknowledgments}
This work was performed under the auspices of the U.S. Department of Energy by Lawrence Livermore National Laboratory under Contract DE-AC52-07NA27344 with IM release number  LLNL-TR-827489. Funding for this work was provided by LLNL Laboratory Directed Research and Development grant 19-SI-004.
This document was prepared as an account of work sponsored by an agency of the United States government. Neither the United States government nor Lawrence Livermore National Security, LLC, nor any of their employees makes any warranty, expressed or implied, or assumes any legal liability or responsibility for the accuracy, completeness, or usefulness of any information, apparatus, product, or process disclosed, or represents that its use would not infringe privately owned rights. Reference herein to any specific commercial product, process, or service by trade name, trademark, manufacturer, or otherwise does not necessarily constitute or imply its endorsement, recommendation, or
favoring by the United States government or Lawrence Livermore National Security, LLC. The views and opinions of authors expressed herein do not necessarily state or reflect those of the United States government or Lawrence Livermore National Security, LLC, and shall not be used for advertising or product endorsement purposes.

\bibliographystyle{siamplain}
\bibliography{/Users/muyskens1/docs/MADSTARE}

\end{document}